\documentclass{article}

\usepackage{PRIMEarxiv}

\usepackage[utf8]{inputenc} % allow utf-8 input
\usepackage[T1]{fontenc}    % use 8-bit T1 fonts
\usepackage{hyperref}       % hyperlinks
\usepackage{url}            % simple URL typesetting
\usepackage{booktabs}       % professional-quality tables
\usepackage{amsmath}        % AMS-style math environments
\usepackage{amsfonts}       % blackboard math symbols
\usepackage{microtype}      % microtypography
\usepackage{fancyhdr}       % header
\usepackage{graphicx}       % graphics
\graphicspath{{media/}}     % organize your images and other figures under media/ folder

\usepackage{listings} % For code listings
\usepackage{color}    % For syntax highlighting
\usepackage[table]{xcolor}

\definecolor{codegray}{gray}{0.9}
\definecolor{darkgreen}{rgb}{0,0.5,0}

\lstnewenvironment{pythoncode}[1][]
{
    \lstset{
        language=Python,
        basicstyle=\small\ttfamily,
        commentstyle=\color{darkgreen},
        keywordstyle=\color{blue},
        stringstyle=\color{magenta},
        showstringspaces=false,
        breaklines=true,
        numbers=left,
        numberstyle=\tiny,
        numbersep=3pt,
        xleftmargin=10pt,
        backgroundcolor=\color{codegray}, % Apply the color to the entire background
        frame=tlbr, % adds a frame on top, left, right, and bottom of listings
        framesep=5pt, % padding between frame and text
        framexleftmargin=5pt, % padding on the left side of the frame
        #1 
    }
}{}

\title{Spyx: A Library for Just-In-Time Compiled Optimization of Spiking Neural Networks
}

\author{
  Kade M. Heckel \\
  University of Cambridge \\
  Department of Engineering \\
  Cambridge, UK\\
  \texttt{kmh70@cam.ac.uk} \\
   \And
  Thomas Nowotny \\
  Sussex AI \\
  School of Engineering and Informatics \\
  University of Sussex \\
  Brighton, UK\\
  \texttt{t.nowotny@sussex.ac.uk} \\
}

\begin{document}
\maketitle

\begin{abstract}

As the role of artificial intelligence becomes increasingly pivotal in modern society, the efficient training and deployment of deep neural networks have emerged as critical areas of focus. Recent advancements in attention-based large neural architectures have spurred the development of AI accelerators, facilitating the training of extensive, multi-billion parameter models. Despite their effectiveness, these powerful networks often incur high execution costs in production environments. Neuromorphic computing, inspired by biological neural processes, offers a promising alternative. By utilizing temporally-sparse computations, Spiking Neural Networks (SNNs) offer to enhance energy efficiency through a reduced and low-power hardware footprint. However, the training of SNNs can be challenging due to their recurrent nature which cannot as easily leverage the massive parallelism of modern AI accelerators. To facilitate the investigation of SNN architectures and dynamics researchers have sought to bridge Python-based deep learning frameworks such as PyTorch or TensorFlow with custom-implemented compute kernels. This paper introduces Spyx, a new and lightweight SNN simulation and optimization library designed in JAX. By pre-staging data in the expansive vRAM of contemporary accelerators and employing extensive JIT compilation, Spyx allows for SNN optimization to be executed as a unified, low-level program on NVIDIA GPUs or Google TPUs. This approach achieves optimal hardware utilization, surpassing the performance of many existing SNN training frameworks while maintaining considerable flexibility. 
\end{abstract}

% keywords can be removed

\section{Introduction}

Since the avalanche of research into transformer neural network architectures resulting from the publication of "Attention is All You Need" \cite{attentionisallyouneed}, the capacity and capability of modern AI accelerators has exploded. A direct result of this is a self-reinforcing cycle deemed the "Hardware Lottery"\cite{hooker2020hardware}, where the demand to train increasingly massive deep neural networks drives tremendous investment in AI hardware which prioritizes parallel and dense matrix computations which then facilitate even greater performance of DNN models. A significant area of concern with the current trajectory of deep learning research and industrial adoption lies in its equally dramatic energy and environmental cost for training and deploying such models\cite{strubell2019energy}. Taking inspiration from biology, the field of neuromorphic computing and the investigation of Spiking Neural Network architectures aim to drastically reduce the computational resources required of ML tools by leveraging sparsity and event-based computation executed on low-power custom hardware. 

The highly heterogeneous architectures and experimental nature of the neuromorphic hardware ecosystem poses unique challenges to developing SNN models, since on-chip learning for one platform may be difficult or impossible to transfer to other systems. Consequently, a number of neuromorphic frameworks developed in the last five years have instead opted to leverage the immense support for GPU-based development to enable the training and simulation of platform-agnostic SNNs on high-performance hardware before conversion and transfer to a specific target device. 

While the motivation of high-power training for low-power underpins almost all SNN training frameworks, each library pursues this objective in a different fashion with associated advantages and drawbacks. In the following we will focus on libraries for direct training of SNNs; the other popular approach of converting a trained artificial neural network to an SNN also adheres to the high-to-low power paradigm but is not a focus of this paper. Some libraries use custom low-level code; for example, mlGeNN \cite{mlgenn} is built upon custom low-level CUDA kernels to implement their neuron dynamics and SpikingJelly \cite{spikingjelly} builds on PyTorch but augments it with custom CUDA, improving their computational speed with some tradeoffs in flexibility and interoperability. In contrast, higher-level frameworks such as snnTorch\cite{eshraghian2021training} and Norse\cite{norse2021} promote flexibility by extending the PyTorch deep learning framework without adding low-level code, enabling a simpler programming experience at the cost of hardware optimization due to dynamic computation graphs though this is a less significant drawback when dealing with multi-node training or if the dataset in question is extremely large. Finally, some libraries such as Lava-dl\cite{lavaDL2021github} and Rockpool\cite{rockpool} offer tooling, features, and abstraction models targeted for their specific neuromorphic computing platforms which are Intel's Loihi \cite{loihi} and SynSense's Xylo \cite{rockpool} hardware families respectively.

The year 2023 has seen the emergence of SNN research powered by the new and growing JAX\cite{jax2018github} high-performance array computing framework developed by Google DeepMind. JAX features Just-In-Time (JIT) compilation for execution on AI accelerators, composable function transformations like vectorization, and automatic differentiation capabilities. Because of its Numpy-like interface and adherence to functional programming paradigms, JAX is host to a growing community of easily connectable libraries such as a differentiable physics simulator\cite{brax2021github}, reinforcement learning environments\cite{gymnax2022github, bonnet2023jumanji, koyamada2023pgx}, or evolutionary optimization tools\cite{evosax2022github}.

Continuing with the trend each approach again tackles slightly different issues in training SNNs, this time capitalizing on the unique features of JAX. The work of Finkbeiner et. al.\cite{finkbeiner2023manycoreIPU} compared the performance of sparse vector computation on NVIDIA GPUs against Graphcore Intelligence Processing Units (IPUs) using a JAX-based implementation, finding significant acceleration for sparse SNN simulation when using the IPU's Multiple-Instruction, Multiple-Data (MIMD) architecture instead of the GPU's Single-Instruction, Multiple-Data (SIMD) construction. Jaxsnn\cite{muller2024jaxsnn}, a continuous-time library for SNN simulation also debuted, drawing parallels to ml-GeN\cite{mlgenn} and Norse\cite{norse2021} by avoiding the discretization of spiking data into a fixed time grid as is common in other SNN libraries. Instead, Jaxsnn is tailored for event-based training and facilitates data-sparse training approaches including time-to-first-spike\cite{ttfs} and EventProp\cite{Wunderlich_2021} and includes support for hardware integration with the BrainScaleS-2 platform\cite{bs2}. 

Concurrent with other JAX-based SNN work and inspired by the approach of snnTorch\cite{eshraghian2021training},  the Spyx\cite{spyx} library was developed to facilitate the efficient investigation of modern evolutionary strategies for optimizing SNNs\cite{heckel_NE4SNNs}.  Spyx combines the flexibility advantages of PyTorch-based frameworks with the performance and efficiency of libraries backed by low-level kernel implementations. Specifically, Spyx allows the user to specify their own surrogate gradient functions or neuron models in only a few lines of code, while the entire training and evaluation loop can be reliably traced and executed on either a GPU or TPU with maximum utilization. % add more motivation for why a new library was necessary

This paper presents a comparison of simulation performance between Spyx and other notable SNN libraries which sit at the intersection of deep learning and neuromorphic computing. The source code is located on GitHub and the package is regularly released and updated on the Python Package Index (PyPI) \cite{kade_heckel_2024_10698926}.

\section{Methods}

\subsection{Library Design}

The development of another SNN library was driven by the desire to exploit the potential of JIT compilation in the JAX ecosystem for accelerating SNN simulations. An important secondary goal of developing Spyx was to minimize the introduction of unfamiliar concepts to users already comfortable with PyTorch-based libraries; this is accomplished through mirroring the general design patterns found in the snnTorch\cite{eshraghian2021training} library for treating SNNs as a special case of recurrent neural networks. We are using DeepMind's Haiku library since it provides an intuitive set of tools for defining recurrent neural networks in an object-oriented fashion before converting them to an efficient and functionally-pure equivalent form suitable for JIT compilation; this makes the learning curve for new users less steep and reduces the codebase footprint. Besides offering object-oriented to functional paradigm conversion, Haiku also has convenient features for unrolling RNN computations and full support for mixed precision training, further increasing hardware utilization. An additional benefit of extending Haiku is that Spyx is highly modular and can seamlessly interface with other libraries in the JAX ecosystem while being executed completely on GPU. For example, one could use Evolutionary Strategies algorithms from the Evosax\cite{evosax2022github} library to evolve populations of parameters of SNNs simulated in Spyx for solving the classic "CartPole" control task implemented in the Gymnax\cite{gymnax2022github} library, with the whole pipeline retaining JIT compilation compatibility. 

Spyx is designed to maximize JIT compilation opportunities and adheres to the same functional programming paradigm JAX follows to enable much of its powerful capabilities. Due to this, Spyx avoids defining classes except when necessary for the implementation of neuron models using the Haiku library; this is done to avoid the possibility of introducing bugs resulting from the mixture of object-oriented programming with the function-oriented rules imposed by function tracing for JIT compilation. Unexpected behavior can occur when compiling a method of a class that incorporates multiple class members. For example, if an "adaptive surrogate gradient" was implemented as a class with a member variable $k$ that parametrizes the slope of the surrogate function as illustrated in Listing \ref{bad-code} and the user were to JIT compile code which called the adaptive\_spike's compute\_gradient function, the evaluation would always use the value of $k=2$. 

\begin{pythoncode}[caption=Example bug from mixing OOP \& FP, label=bad-code]
class adaptive_spike():
    def __init__(k=1):
        self.k = k

    def compute_gradient(x):
        self.k = self.k + 1
        return 1 / (1 + k*jnp.abs(x))**2
\end{pythoncode}

To avoid problems of this kind, Spyx uses higher-order-functions that take hyperparameters of a function such as the steepness of the surrogate gradient and returns the compiled function. This design decision reinforces to the user that many of the features in Spyx are functionally pure and stateless, and if they want values in a computation to change over time they need to make them arguments of the function. The rest of this section will explain the design principles of Spyx, building up the different components necessary to efficiently train SNNs in JAX.

\subsubsection{spyx.axn: Surrogate Gradient Functions}

Spyx offers a variety of surrogate gradient functions that are common throughout SNN optimization literature, such as SuperSpike\cite{superspike}, Arctangent, and Boxcar\cite{boxcar}, among others. Defining custom gradient calculations in JAX is quite similar to how it is performed in PyTorch; all that is required for creating surrogate gradients in Spyx is to define a function which returns the value of the forward computation and another function which computes the gradient. This is detailed in Listing \ref{customsurrogate-code} which shows the custom surrogate gradient builder function in Spyx. The default surrogate gradient is a straight-through estimator with a Heaviside activation which users can flexibly using their own functions as desired. This simple modular approach empowers the user to try different combinations of forward and backward functions including the use of ternary or other low-precision activations.

\begin{pythoncode}[caption=Constructing Spiking Activation Functions in JAX, label=customsurrogate-code]
import jax.numpy as jnp

def custom(bwd=lambda x: x,  fwd=lambda x: heaviside(x)):

    @jax.custom_gradient
    def f(x):
        return fwd(x), lambda g: g * bwd(x)

    return jax.jit(f)
\end{pythoncode}

The popular surrogate gradient functions included in Spyx are offered via higher-order-functions which define the backwards pass and then call the custom surrogate constructor. As it can be seen, surrogate gradients can be defined in an extremely compact manner, making it easy for the user to create their own and for the community to contribute new methods over time.
\newpage
\begin{pythoncode}[caption=Defining a Surrogate Gradient, label=surrogate-code]
def superspike(k=25):
    def grad_superspike(x):
        return 1 / (1 + k*jax.numpy.abs(x))**2
    
    return custom(grad_superspike, heaviside)

def triangular(k=0.5):
    def grad_traingle(x):
        return jax.numpy.maximum(0, 1-jax.numpy.abs(k*x))
    
    return custom(grad_traingle, heaviside)
\end{pythoncode}

\subsubsection{spyx.nn: Neuron Models}

Neuron models including Leaky-Integrate-and-Fire and other variations including non-spiking, non-leaking, and neurons with adaptive firing thresholds are implemented as extensions of RNN cores in DeepMind's Haiku library. Listing \ref{lif-code} showcases the implementation of a LIF neuron in Spyx; it can be seen as a direct mirroring of the dynamics encoded by the snnTorch library adhering to the following equations:

\begin{align}
S_t &= \phi\left(V_t-\theta\right) \\
V_{t+1} &= \beta V_t + X - S_t\theta
\end{align}
where $S_t$ denotes the spikes for the current iteration $t$, $V_t$ the neuron membrane potential, $\theta$ the firing threshold, $\beta$ the inverse time constant or the voltage decay rate $\beta=\exp(-\Delta t/\tau_m)$, and $X$ the neuron inputs; the weighting of different inputs is offloaded to the preceding use of a haiku.Linear layer to build off existing tools and minimize redundancy. When training with surrogate gradient methods, the activation function $\phi$ is the Heaviside Step Function whose gradient is replaced by a suitable finite surrogate gradient function.

\begin{pythoncode}[caption=A Simplified Leaky-Integrate-and-Fire Neuron Implementation, label=lif-code]
class LIF(haiku.RNNCore):
    def __init__(self, hidden_shape: tuple, threshold=1, activation = superspike(), name="LIF"):
        super().__init__(name=name)
        self.hidden_shape = hidden_shape
        self.act = activation
    
    def __call__(self, x, V):
        beta = haiku.get_parameter("beta", self.hidden_shape,
                            init=haiku.initializers.TruncatedNormal(0.25, 0.5))
        beta = jax.numpy.clip(beta, 0, 1)
            
        # calculate whether spike is generated, and update membrane potential
        spikes = self.act(V - 1)
        V = beta*V + x - spikes
        
        return spikes, V

    def initial_state(self, batch_size): 
        return jax.numpy.zeros((batch_size,) + self.hidden_shape)
\end{pythoncode}
Spyx enables researchers to quickly iterate on their ideas, permitting the user to define the model dynamics in terms of Numpy-like calculations while still achieving speeds comparable to writing a low-level CUDA kernel in C/C++. With the spiking neuron dynamics defined as an extension of an RNN module, the user can leverage the machinery in the Haiku library to quickly build an SNN as shown in Listing \ref{net-code}. The sequential stacking of layers without worrying about explicitly managing the recurrent state of the LIF layers is a major advantage of building on top of the Haiku library instead of other JAX-based frameworks which don't offer the same object-oriented to functional conversion tools.

\newpage
\begin{pythoncode}[caption=A simple SNN for the SHD dataset, label=net-code]
def shd_snn(x): 
    surrogate = spyx.axn.arctan()
    core = hk.DeepRNN([
        hk.Linear(64, with_bias=False),
        spyx.nn.LIF((64,), activation=surrogate),
        hk.Linear(64, with_bias=False),
        spyx.nn.LIF((64,), activation=surrogate),
        hk.Linear(20, with_bias=False),
        spyx.nn.LI((20,))
    ])
    spikes, V = hk.dynamic_unroll(core, x, core.initial_state(x.shape[0]), time_major=False)
    
    return spikes, V

key = jax.random.PRNGKey(0)
SNN = hk.without_apply_rng(hk.transform(shd_snn))
params = SNN.init(rng=key, x=sample_batch)
\end{pythoncode}

After transforming the model function and initializing the network, it is now a side-effect free function that can be JIT compiled inside of a training loop. Additionally, Haiku offers the ability to statically unroll RNNs and do batch-parallel application of stateless computations such as linear and convolutional layers; this enables Spyx to simulate SNNs in a similar fashion as the high-performance SpikingJelly library's multistep mode. The underlying implementation of RNNs/SNNs in Haiku and Spyx relies on JAX's scan operator, which dispatches a CUDA kernel for each iteration of the loop. By increasing the size of the dynamic unroll or using a static unroll, the user can fuse together a variable number of computations to reduce overhead incurred by CPU-GPU communication. This layer-by-layer method of computation is akin to the multi-step mode used in in the SpikingJelly library\cite{spikingjelly}.

\subsubsection{spyx.fn: Loss and Regularization Functions}

Spyx currently includes loss functions for rate coding objectives, such as integral cross-entropy and spike rate mean squared error (MSE). To evaluate the loss and gradient of an SNN in Spyx, the user defines a function which takes the network parameters, the data, and the labels and returns the value computed by the loss function. This user-defined function can then be converted using JAX's value\_and\_grad() transformation into a function which returns the loss and the network's gradient for use with optimizers from the Optax library\cite{optax2020github}.

\begin{pythoncode}[caption=Evaluating the loss and gradient of an SNN, label=net_eval-code]
Loss =  spyx.fn.integral_crossentropy()

@jax.jit
def net_eval(weights, events, targets):
    readout = SNN.apply(weights, events)
    traces, V_f = readout
    return Loss(traces, targets)

loss_and_surrogate_grad = jax.value_and_grad(net_eval) 
\end{pythoncode}

Regularizing the activity of intermediate layers in Spyx is supported both for avoiding dead neurons and for tempering layer activity to discourage high firing rates. This is accomplished through adding ActivityRegularization layers to the network definition after spiking neuron layers and transforming the network definition with state; this causes the network's apply() function to additionally return a JAX PyTree containing the spike counts for each neuron per example in the mini-batch. This dictionary-like data structure can then be fed into a regularization function that computes and returns a scalar penalty to include in the network evaluation function alongside the supervised training loss. For code demonstrating this process please refer to the library's repository on GitHub \cite{spyx}.

\subsubsection{spyx.data: Data Augmentation and Dynamic Decompression}

To optimize GPU usage, several optimizations must also be made to the handling of data. To enable full JIT compilation and on-accelerator training, the full data needs to be loaded into the GPU vRAM and the operations on the data need to be traceable by JAX; the incentive to do so is that streaming data from the CPU induces I/O latency thereby reducing training throughput. The user is free to use any loading and preprocessing pipeline to read their data into CPU memory. From there, the dataset should be converted to a single Numpy array organized with the shape $[\#Batches, Batch Size, Time, ...]$. Calling the numpy.packbits() function with the temporal axis specified compresses the dataset by 87.5\% prior to its transfer to the GPU. The compressed data is then used directly with dynamic decompression during simulations (see Listing \ref{data-code} below). Spyx offers data-specific utility functions for randomly shifting inputs along the specified data axes as well as shuffling batches inside of JIT compiled training loops without needing to involve the CPU-based dataloader utilities as shown in Listings \ref{augment-code} and \ref{shuffle-code}.

\begin{pythoncode}[caption=On-accelerator data augmentation in Spyx, label=augment-code]
def shift_augment(max_shift=10, axes=(-1,)):
    def _shift(data, rng):
        shift = jax.random.randint(rng, (len(axes),), -max_shift, max_shift)
        return jnp.roll(data, shift, axes)
    
    return jax.jit(_shift)
\end{pythoncode}

The shuffling function works by taking the full dataset stored as a JAX.numpy array and permutes it before reshaping it into mini-batches. By doing this processing on the GPU, significant time is saved compared to using a CPU-based dataloader when working on systems without unified memory. 

\begin{pythoncode}[caption=Efficiently shuffling a dataset on GPU, label=shuffle-code]
def shuffler(dataset, batch_size):
    x, y = dataset
    cutoff = (y.shape[0] // batch_size) * batch_size
    data_shape = (-1, batch_size) + x.shape[1:]

    def _shuffle(dataset, shuffle_rng):
        x, y = dataset

        indices = jax.random.permutation(shuffle_rng, y.shape[0])[:cutoff]
        obs, labels = x[indices], y[indices]

        obs = jnp.reshape(obs, data_shape)
        labels = jnp.reshape(labels, (-1, batch_size)) 

        return (obs, labels)
    return jax.jit(_shuffle)
\end{pythoncode}

The compressed input spike data can be unpacked with functions offered by JAX for packing and unpacking bits to and from unsigned 8-bit integer arrays. This is an advantage unique to a JAX-based approach since PyTorch-based frameworks lack the same underlying pack/unpack operations on tensors. This diminishes a major drawback of time-based dense computation SNN simulators: the fact that a majority of the input data is wasted storing zeros. By storing 8 spikes per byte and only materializing the binary form of the data batch-by-batch, Spyx achieves significant memory efficiencies over other frameworks when working with neuromorphic data. The decompression of the spike data is shown on line 5 of Listing \ref{data-code}, which shows how the gradient computation and optimizer updates can be fused together as a single JIT compiled function in Spyx. 

\begin{pythoncode}[caption=Dynamically decompressing spiking data, label=data-code]
    @jax.jit
    def train_step(state, data):
        grad_params, opt_state = state
        events, targets = data
        events = jnp.unpackbits(events, axis=1) # decompress temporal axis
        # compute loss and gradient
        loss, grads = surrogate_grad(grad_params, events, targets)
        # generate updates based on the gradients and optimizer
        updates, opt_state = opt.update(grads, opt_state, grad_params)
        # return the updated parameters
        new_state = [optax.apply_updates(grad_params, updates), opt_state]
        return new_state, loss
\end{pythoncode}

\subsubsection{spyx.nir: Import and Export SNNs}

To achieve the aim of training rapidly on high-performance AI accelerators and then deploying to energy-efficient neuromorphic hardware, Spyx implements an interface to the newly introduced Neuromorphic Intermediate Representation \cite{NIR2023} as an easy way to serialize and save trained models. By creating a common format for describing SNNs through their dynamics, the NIR standard allows for the translation of models to or from any supporting software framework to a number of supporting hardware platforms. For exporting an SNN from Spyx, the PyTree data structure which is a dictionary of layer names and associated parameters is iterated over to produce a NIR Graph. This graph contains two data structures: a list of NIR nodes containing re-scaled weight matrices and the appropriate spiking neuron layer parameters and an edge list specifying the connectivity between each of the layers. The need to re-scale weights is a result of needing to generalize from the simplifying assumptions in the formulation of spiking neurons Spyx such as the size of timesteps taken during network simulation; for more details see the NIR paper\cite{NIR2023}. To load graphs from NIR into Spyx requires preprocessing as well to re-scale the weights and reorder the graph edges, after which a Spyx SNN is constructed and its parameters loaded in.

\begin{pythoncode}[caption=NIR, label=nir-code]
import spyx
import nir

# Exporting a Spyx network to NIR:
export_params = spyx.nir.reorder_layers(initial_params, optimized_params)
Graph = spyx.nir.to_nir(export_params, input_shape, output_shape)
nir.write("./spyx_network.nir", Graph)

# Importing an SNN to Spyx:
NIR_graph = nir.read("SNN.nir")
SNN, params = spyx.nir.from_nir(NIR_graph, input_data, dt=1)
\end{pythoncode}

\section{Experiments \& Results}
\label{sec:results}

\subsection{Experimental Setup}
The fidelity of models trained and simulated in the Spyx framework is supported by the results presented in \cite{NIR2023}, where explicitly recurrent and convolutional models trained in other frameworks were imported to Spyx and executed with near equivalent dynamics. To compare the training performance of Spyx against other popular libraries, we selected the SHD\cite{SHD} and NMNIST\cite{NMNIST} datasets as benchmarks and evaluated the time needed to train feed-forward and convolutional SNN architectures on the respective datasets. We use snnTorch\cite{eshraghian2021training}, SpikingJelly\cite{spikingjelly}, and mlGeNN\cite{mlgenn} for comparison since snnTorch inspired the formulation of the Spyx neuron models and SpikingJelly and mlGeNN are popular modern frameworks for event-based machine learning. The ability to compile SNNs constructed in snnTorch and other PyTorch-based libraries is experimental as of the time of writing with support limited to LIF neurons; this is in contrast to Spyx which inherently supports JIT compilation for all of its neuron models as well as the training loops themselves. Experiments were conducted on an NVIDIA A6000 which has 48GB of vRAM. While some of the libraries have the ability to utilize mixed-precision training to accelerate computation, we used 32-bit floating point across all frameworks to provide a fair comparison. With the exception of mlGeNN we staged all training data in VRAM as a single large tensor and iterated over it in order to eliminate I/O latency between the CPU and GPU.

\subsection{Spiking Heidelberg Digits Benchmark}
We constructed equivalent feed-forward SNN architectures in Spyx, snnTorch, and mlGeNN and trained them for 100 epochs each. We did not use any additional training procedures such as data augmentation in order to make the experiments as informative about the core neural network simulation as possible. Accuracy values for each network were on the order of 70\%-75\%. The actual scores are not a focus of the benchmark so long as they show that a reasonable model was trained.

We used 256 time steps when rasterizing the data to keep experiment times reasonable while still presenting a legitimate test to the frameworks. We trained each SNN with 1 warm-up run conducted before 5 measured trials for frameworks relying on JIT compilation. Table \ref{tab:shd} contains the observed mean wall-clock times and standard deviations. SNNs trained in Spyx were compiled with the Haiku feature for unrolling RNN loops, with 32 steps being unrolled; this results in a noticeable performance boost relative to executing one step at a time since each iteration of the scan operation used in implementing efficient JIT compiled loops in JAX launches a CUDA kernel. The larger a scan operation in JAX is unrolled, the less communication overhead between the CPU and GPU is incurred at the expense of a longer compile time. The uncompiled snnTorch model was only evaluated for a network with hidden layer size of 128 neurons because the training times are substantially larger than for the other frameworks. Additionally, we repeated the uncompiled snnTorch-128 experiments only 3 times compared to 5 times for the other frameworks for the same reason. While mlGeNN was observed to achieve high GPU utilization percentages during training, the discrepancy in training times compared to the other frameworks can be attributed to the fact that the mlGeNN dataloading pipeline is based on the CPU, which introduces additional latency into the training process.

\begin{table}[t]
 \caption{Timing (in seconds) benchmark for training a feed-forward SNN for 100 epochs on the SHD dataset.}
    \label{tab:shd}
    \centering
    \begin{tabular}{lccc} 
    \toprule  
 Batch Size:& 64&128&256\\ \midrule 
          snnTorch-128&4543.3$\pm$41.0&   2313.2$\pm$12.0&1212.03$\pm$ 0.9\\ 
 snnTorch-128 Compiled& 195.6 $\pm$ 1.2& 103.3 $\pm$ 1.0&58.8 $\pm$ 0.1\\
 Spyx-128 (ours)& \textbf{69.3$\pm$0.1}& \textbf{46.0$\pm$0.1}&\textbf{35.5$\pm$0.1}\\  
          mlGeNN-128&215.7$\pm$1.6& 
     160.8$\pm$1.6&123.5$\pm$1.0\\ 
     \midrule
 snnTorch-512 Compiled& 269.9$\pm$6.0& 137.5$\pm$3.8&78.7$\pm$0.6\\ 
 Spyx-512 (ours)& \textbf{98.7$\pm$0.2}& \textbf{80.2$\pm$0.2}&\textbf{64.1$\pm$0.2}\\
 mlGeNN-512& 690.9$\pm$84.9& 418.4$\pm$39.4&339.5$\pm$23.7\\
 \bottomrule
 \end{tabular}
\end{table}

\subsection{Neuromorphic MNIST Benchmark}

For the second set of benchmarks, we trained convolutional SNNs on the Neuromorphic-MNIST dataset \cite{NMNIST}. This dataset consists of 60,000 images of handwritten digits that were captured by an event camera performing triangular saccades. NMNIST is -- like SHD -- a standard dataset for neuromorphic research. Since the EventProp implementation for Convolutional SNNs is still experimental in mlGeNN, we replaced it in this benchmark with SpikingJelly, which also offers custom CUDA kernel acceleration through its high-performance CuPy backend option. We trained the CSNNs for 20 epochs each on a random subsample of half of the standard training dataset. Each of the 30,000 images was rasterized into 64 time steps at the original 34 by 34 resolution. We used a small network as described in the snnTorch Tutorial \cite{eshraghian2021training} (denoted 12C5-MP2-32C5-MP2-800FC10):
\begin{itemize}
    \item 12C5 is a 5 × 5 convolutional kernel with 12 filters
    \item MP2 is a 2 × 2 max-pooling function
    \item 800FC10 is a fully-connected layer that maps 800 neurons to 10 output neurons
\end{itemize}

In the second set of runs we used a larger network with twice the number of filters in each convolutional layer, resulting in a CSNN with structure 24C5-MP2-64C5-MP2-1600FC10. In terms of neuron models, we benchmarked using the SpikingJelly ParametricLIFNode and compared against the Spyx LIF neuron with beta specified to be a learnable scalar, which ensures equal parameter counts and computational demand for both frameworks. The benchmark experiment notebooks are located in the research/paper subdirectory of the Spyx GitHub Repository \url{https://github.com/kmheckel/spyx}.

The observed wall clock times for the NMNIST benchmark are shown in table \ref{tab:nmnist}. 
SpikingJelly can achieve impressive performance when its CuPy backend is used and SNN model is executed in multistep mode. In this setting, stateless transformations such as linear or convolutional layers are applied to the entire batch in parallel and the recurrent neuron model dynamics are unrolled and executed in a layer-by-layer fashion. It is easy to construct SNNs in the same fashion in Spyx which yields similarly dramatic improvements in training speed. The runtimes shown for Spyx in table \ref{tab:nmnist} are observed after performing ahead-of-time compilation which took on average approximately 30-40 seconds; the compilation time for models built in Spyx is linearly related to the number of time steps in the data when performing a static unroll of the SNN while dynamic unrolls have constant compile times with respect to the temporal resolution. We evaluated each configuration 5 times with one warm-up trial to mitigate effects of initial compilation and caching. Impressively, Spyx is at worst within 5\% of the training speed of the SpikingJelly implementation while not needing to rely on any custom CUDA code generation; its performance is achieved purely through the use of JAX's JIT compilation to XLA\cite{XLA}.

\begin{table}[t]
  \caption{Timing (in Seconds) benchmark for training a convolutional SNN on the NMNIST dataset.}
    \label{tab:nmnist}
    \centering
    \begin{tabular}{lrrr} \toprule
 Batch Size:& 32&64&128\\ \midrule
 Spyx (S) (ours)& 135$\pm$0.1& 128.4$\pm$0.1&127.3$\pm$0.0\\ 
          SpikingJelly (S)&129.4$\pm$0.4& 
     126.5 $\pm$ 0.2&125.3$\pm$ 0.1\\ \midrule
 Spyx (L) (ours)& 247.2$\pm$0.1& 241.2$\pm$0.1&236.5$\pm$0.0\\
 SpikingJelly (L)& 234.0 $\pm$ 0.1& 229.7 $\pm$ 0.1&227.6 $\pm$ 0.1\\
 \bottomrule
 \end{tabular}
   
\end{table}

\section{Discussion}

In this work, we have presented Spyx, a JAX-based system for spiking neural network simulation and learning, and benchmarked it against a number of popular frameworks for spike-based machine learning. We observed a competitive runtime performance even in comparison to frameworks that use custom CUDA code generation for runtime optimisation.

% talk about benchmarks
It is important to note that benchmarks by practical necessity cannot be fully comprehensive and must focus on a useful subset of relevant operation regimes. For instance, here we have focused on a medium to low temporal resolution regime in which dense computation frameworks such as Spyx and snnTorch perform well. In contrast, the sparse data structures underlying mlGeNN's functionality are better suited to high-temporal resolution regimes, where the frameworks relying on dense computation structures begin to suffer due to increased data sparsity. Given these notable differences between mlGeNN and the other frameworks, the comparison is not as direct as between Spyx and the other libraries that were evaluated. 

Furthermore, the models trialled in this paper are of relatively small size relative to the larger deep-learning-inspired architectures being probed by some in the neuromorphic research community. However, it is reasonable to assume that the observed differences in the runtime of the smaller models offer some information on the relative efficiency of the frameworks for other model sizes as long as operating regime in terms of sufficient GPU memory and cache usage patterns does not change.

% talk about torch JIT being fickle
While PyTorch recently introduced additional JIT compilation abilities in version 2.0, it is still a work in progress and not as mature as the features found in JAX. Despite the experimental nature of the support for JIT compilation by PyTorch-based SNN libraries, the results from our benchmarks show promise that libraries such as snnTorch\cite{eshraghian2021training} and Norse\cite{norse2021} will also benefit substantially in the future as the PyTorch JIT capabilities improve. In the meantime, Spyx offers a consistent and reliable method to perform high-speed optimization of SNN architectures while maintaining an easy API for researchers.

% discuss and highlight Spyx advantages
Since the entire training loop  in Spyx is JIT-compiled, the Python interpreter is able to leverage JAX's asynchronous dispatch and execution to "run ahead" of the GPU and ensure it is always at maximum utilization. This is especially beneficial when working with other JAX-based libraries since Python can be removed from the critical path. For example, Spyx could be used in conjunction with Brax, a fully-differentiable and JIT-compilable physics simulator implemented in JAX, to rapidly experiment with SNNS learning or evolving to solve embodied tasks such as locomotion. The growing JAX ecosystem offers exciting opportunities for investigating neuromorphic control systems and other experiments on SNNs with methods such as neuroevolution and Bayesian optimization at speeds and scales previously unattainable. Additionally, the compact structure and minimal dependencies of Spyx make it extremely easy to install and highly reliable. Another advantage conferred by building within the JAX ecosystem is that users only have to write a few lines of Python code to define a custom neuron model or surrogate gradient and still get to retain training and evaluation speeds on par with that of libraries which utilize lower level interfaces with CUDA. 

% note drawbacks or tradeoffs with Spyx
It is worth noting that the extreme speedups in Spyx are the result of minimizing work done on the CPU and by the Python interpreter; when dealing with datasets that exceed GPU vRAM capacity such as hundreds of gigabytes of autonomous vehicle sensor data or when training a large enough model to require multi-GPU computation, the blocking induced by inter-process communications and data transfer will incur overhead. The highly-compiled nature of Spyx also introduces a key-rigidity: it is not amenable to scenarios involving ragged and variable shape tensors. This fixed-shape poses a disadvantage if events have highly variable timescales since short events will require excessive padding. It also makes methods involving the dynamic growth and modification of the network architecture impractical and inefficient since each change would require recompilation. 

% future directions.
Future enhancements to the Spyx library will focus on incorporating features proposed by the research community including new neuron models and predefined building blocks for larger SNN architectures. For example, the addition of tooling to support latency-based spike coding and Phasor network\cite{deepphasornet, spikingphasornetwork} training through JAX's complex-valued auto-differentiation is a possible direction. Additionally, the inclusion of stochastic spiking mechansims such as Stochastic Parallelizable Spiking Neurons\cite{SPSN} to facilitate the exploration of spiking state space models\cite{stan2023SpikingSSMs} poses exciting opportunities to train SNNs at high throughput through the use of Fourier transforms and parallel scans to replace recurrent computations. Finally, the performance of the library could be improved even further through the definition of custom kernels using Pallas, which is the kernel language for JAX and permits more granular control of GPU and TPU hardware\cite{jax2018github}.

\section{Conclusion}

The experiments conducted in this work demonstrate the potential value of Spyx for accelerating SNN research. By effectively leveraging Just-In-Time compilation, Spyx is almost able to match the training performance of libraries which leverage custom CUDA implementations while retaining the flexibility and interoperability seen in PyTorch-based libraries. With its compact footprint and streamlined design, Spyx offers researchers the ability to rapidly iterate on innovative concepts without needing to re-implement lower level primitives to achieve acceleration. Finally, given the ever-expanding ecosystem of libraries and tools built in JAX, Spyx is poised to enable exciting new directions of rapid research at the intersection of spiking neural networks and deep learning.

\section*{Acknowledgments}
This work was supported by funding from the Marshall Aid Commemoration Commission and by the EPSRC (grant EP/S030964/1). We would like to thank Steven Abreu and Gregor Lenz for their help with interfacing Spyx to NIR and improving the documentation, respectively.

\section{Code Availability}
Spyx is available publicly on Github at \url{https://github.com/kmheckel/spyx} under the MIT licence.

%Bibliography
\bibliographystyle{unsrt}  
\bibliography{references}  

\begin{thebibliography}{10}

\bibitem{attentionisallyouneed}
Ashish Vaswani, Noam Shazeer, Niki Parmar, Jakob Uszkoreit, Llion Jones,
  Aidan~N Gomez, \L~ukasz Kaiser, and Illia Polosukhin.
\newblock Attention is all you need.
\newblock In I.~Guyon, U.~Von Luxburg, S.~Bengio, H.~Wallach, R.~Fergus,
  S.~Vishwanathan, and R.~Garnett, editors, {\em Advances in Neural Information
  Processing Systems}, volume~30. Curran Associates, Inc., 2017.

\bibitem{hooker2020hardware}
Sara Hooker.
\newblock The hardware lottery.
\newblock {\em CoRR}, abs/2009.06489, 2020.

\bibitem{strubell2019energy}
Emma Strubell, Ananya Ganesh, and Andrew McCallum.
\newblock Energy and policy considerations for deep learning in nlp, 2019.

\bibitem{mlgenn}
James~Paul Turner, James~C Knight, Ajay Subramanian, and Thomas Nowotny.
\newblock mlgenn: accelerating snn inference using gpu-enabled neural networks.
\newblock {\em Neuromorphic Computing and Engineering}, 2(2):024002, mar 2022.

\bibitem{spikingjelly}
Wei Fang, Yanqi Chen, Jianhao Ding, Zhaofei Yu, Timothée Masquelier, Ding
  Chen, Liwei Huang, Huihui Zhou, Guoqi Li, and Yonghong Tian.
\newblock Spikingjelly: An open-source machine learning infrastructure platform
  for spike-based intelligence.
\newblock {\em Science Advances}, 9(40):eadi1480, 2023.

\bibitem{eshraghian2021training}
Jason~K Eshraghian, Max Ward, Emre Neftci, Xinxin Wang, Gregor Lenz, Girish
  Dwivedi, Mohammed Bennamoun, Doo~Seok Jeong, and Wei~D Lu.
\newblock Training spiking neural networks using lessons from deep learning.
\newblock {\em Proceedings of the IEEE}, 111(9):1016--1054, 2023.

\bibitem{norse2021}
Christian Pehle and Jens~Egholm Pedersen.
\newblock {Norse - A deep learning library for spiking neural networks},
  January 2021.
\newblock Documentation: https://norse.ai/docs/.

\bibitem{lavaDL2021github}
{Lava: A software framework for neuromorphic computing}, 2021.

\bibitem{rockpool}
Dylan~R. Muir, Felix Bauer, and Philipp Weidel.
\newblock Rockpool documentaton, September 2019.

\bibitem{loihi}
Garrick Orchard, E.~Paxon Frady, Daniel Ben~Dayan Rubin, Sophia Sanborn,
  Sumit~Bam Shrestha, Friedrich~T. Sommer, and Mike Davies.
\newblock Efficient neuromorphic signal processing with loihi 2, 2021.

\bibitem{jax2018github}
James Bradbury, Roy Frostig, Peter Hawkins, Matthew~James Johnson, Chris Leary,
  Dougal Maclaurin, George Necula, Adam Paszke, Jake Vander{P}las, Skye
  Wanderman-{M}ilne, and Qiao Zhang.
\newblock {JAX}: composable transformations of {P}ython+{N}um{P}y programs,
  2018.

\bibitem{brax2021github}
C.~Daniel Freeman, Erik Frey, Anton Raichuk, Sertan Girgin, Igor Mordatch, and
  Olivier Bachem.
\newblock Brax - a differentiable physics engine for large scale rigid body
  simulation, 2021.

\bibitem{gymnax2022github}
Robert~Tjarko Lange.
\newblock {gymnax}: A {JAX}-based reinforcement learning environment library,
  2022.

\bibitem{bonnet2023jumanji}
Clément Bonnet, Daniel Luo, Donal Byrne, Shikha Surana, Vincent Coyette, Paul
  Duckworth, Laurence~I. Midgley, Tristan Kalloniatis, Sasha Abramowitz,
  Cemlyn~N. Waters, Andries~P. Smit, Nathan Grinsztajn, Ulrich A.~Mbou Sob,
  Omayma Mahjoub, Elshadai Tegegn, Mohamed~A. Mimouni, Raphael Boige, Ruan
  de~Kock, Daniel Furelos-Blanco, Victor Le, Arnu Pretorius, and Alexandre
  Laterre.
\newblock Jumanji: a diverse suite of scalable reinforcement learning
  environments in jax, 2023.

\bibitem{koyamada2023pgx}
Sotetsu Koyamada, Shinri Okano, Soichiro Nishimori, Yu~Murata, Keigo Habara,
  Haruka Kita, and Shin Ishii.
\newblock Pgx: Hardware-accelerated parallel game simulators for reinforcement
  learning.
\newblock In {\em Advances in Neural Information Processing Systems}, 2023.

\bibitem{evosax2022github}
Robert~Tjarko Lange.
\newblock evosax: Jax-based evolution strategies.
\newblock {\em arXiv preprint arXiv:2212.04180}, 2022.

\bibitem{finkbeiner2023manycoreIPU}
Jan Finkbeiner, Thomas Gmeinder, Mark Pupilli, Alexander Titterton, and Emre
  Neftci.
\newblock Harnessing manycore processors with distributed memory for
  accelerated training of sparse and recurrent models, 2023.

\bibitem{muller2024jaxsnn}
Eric M{\"u}ller, Moritz Althaus, Elias Arnold, Philipp Spilger, Christian
  Pehle, and Johannes Schemmel.
\newblock jaxsnn: Event-driven gradient estimation for analog neuromorphic
  hardware.
\newblock {\em arXiv preprint arXiv:2401.16841}, 2024.

\bibitem{ttfs}
J.~Göltz, L.~Kriener, A.~Baumbach, S.~Billaudelle, O.~Breitwieser, B.~Cramer,
  D.~Dold, A.~F. Kungl, W.~Senn, J.~Schemmel, K.~Meier, and M.~A. Petrovici.
\newblock Fast and energy-efficient neuromorphic deep learning with first-spike
  times.
\newblock {\em Nature Machine Intelligence}, 3(9):823–835, September 2021.

\bibitem{Wunderlich_2021}
Timo~C. Wunderlich and Christian Pehle.
\newblock Event-based backpropagation can compute exact gradients for spiking
  neural networks.
\newblock {\em Scientific Reports}, 11(1), June 2021.

\bibitem{bs2}
Christian Pehle, Sebastian Billaudelle, Benjamin Cramer, Jakob Kaiser,
  Korbinian Schreiber, Yannik Stradmann, Johannes Weis, Aron Leibfried, Eric
  Müller, and Johannes Schemmel.
\newblock The brainscales-2 accelerated neuromorphic system with hybrid
  plasticity.
\newblock {\em Frontiers in Neuroscience}, 16, 2022.

\bibitem{spyx}
Kade Heckel, Steven Abreu, Gregor Lenz, and Thomas Nowotny.
\newblock kmheckel/spyx: v0.1.17, February 2024.

\bibitem{heckel_NE4SNNs}
Kade Heckel.
\newblock Neuroevolution of spiking neural networks, February 2024.

\bibitem{kade_heckel_2024_10698926}
Kade Heckel, Steven Abreu, Gregor Lenz, and Thomas Nowotny.
\newblock kmheckel/spyx: Paper edition, February 2024.

\bibitem{superspike}
Friedemann Zenke and Surya Ganguli.
\newblock Superspike: Supervised learning in multilayer spiking neural
  networks.
\newblock {\em Neural Computation}, 30(6):1514–1541, June 2018.

\bibitem{boxcar}
Jacques Kaiser, Hesham Mostafa, and Emre Neftci.
\newblock Synaptic plasticity dynamics for deep continuous local learning
  (decolle).
\newblock {\em Frontiers in Neuroscience}, 14, 2020.

\bibitem{optax2020github}
Matteo Hessel, David Budden, Fabio Viola, Mihaela Rosca, Eren Sezener, and Tom
  Hennigan.
\newblock Optax: composable gradient transformation and optimisation, in jax!,
  2020.

\bibitem{NIR2023}
Jens~E. Pedersen, Steven Abreu, Matthias Jobst, Gregor Lenz, Vittorio Fra,
  Felix~C. Bauer, Dylan~R. Muir, Peng Zhou, Bernhard Vogginger, Kade Heckel,
  Gianvito Urgese, Sadasivan Shankar, Terrence~C. Stewart, Jason~K. Eshraghian,
  and Sadique Sheik.
\newblock Neuromorphic intermediate representation: A unified instruction set
  for interoperable brain-inspired computing.
\newblock {\em arXiv}, 2023.

\bibitem{SHD}
Benjamin Cramer, Yannik Stradmann, Johannes Schemmel, and Friedemann Zenke.
\newblock Heidelberg spiking datasets, 2019.

\bibitem{NMNIST}
Garrick Orchard, Ajinkya Jayawant, Gregory Cohen, and Nitish~V. Thakor.
\newblock Converting static image datasets to spiking neuromorphic datasets
  using saccades.
\newblock {\em CoRR}, abs/1507.07629, 2015.

\bibitem{XLA}
Amit Sabne.
\newblock Xla : Compiling machine learning for peak performance, 2020.

\bibitem{deepphasornet}
Wilkie Olin-Ammentorp and Maxim Bazhenov.
\newblock Deep phasor networks: Connecting conventional and spiking neural
  networks, 2021.

\bibitem{spikingphasornetwork}
Connor Bybee, E.~Paxon Frady, and Friedrich~T. Sommer.
\newblock Deep learning in spiking phasor neural networks, 2022.

\bibitem{SPSN}
Sidi Yaya~Arnaud Yarga and Sean U.~N. Wood.
\newblock Accelerating snn training with stochastic parallelizable spiking
  neurons.
\newblock In {\em 2023 International Joint Conference on Neural Networks
  (IJCNN)}, pages 1--8, 2023.

\bibitem{stan2023SpikingSSMs}
Matei~Ioan Stan and Oliver Rhodes.
\newblock Learning long sequences in spiking neural networks, 2023.

\end{thebibliography}

\end{document}